%% file: 00_main.tex
\documentclass[10pt,twocolumn,letterpaper]{article}

\usepackage{cvpr}              %
\usepackage[accsupp]{axessibility}
\usepackage{graphicx}
\usepackage{amsmath}
\usepackage{amssymb}
\usepackage{booktabs}
\usepackage{adjustbox}
\usepackage{multirow}
\usepackage{graphics}
\usepackage{subcaption}
\usepackage{multicol}
\usepackage{subcaption}
\usepackage{comment}
\usepackage{color}
\usepackage[table]{xcolor}
\usepackage{float}
\usepackage{dsfont}
\usepackage{placeins}

\usepackage[pagebackref,breaklinks,colorlinks]{hyperref}

\usepackage[capitalize]{cleveref}
\crefname{section}{Sec.}{Secs.}
\Crefname{section}{Section}{Sections}
\Crefname{table}{Table}{Tables}
\crefname{table}{Tab.}{Tabs.}

\makeatletter
\def\blfootnote{\xdef\@thefnmark{}\@footnotetext}
\makeatother

\newcommand{\myparagraph}[1]{\vspace{0.2em}\noindent\textbf{#1}}
\newcommand{\boldparagraph}[1]{\vspace{0.3em}\noindent{\bf #1} }
\newcommand{\norm}[1]{\left\lVert#1\right\rVert}

\def\cvprPaperID{4351} %
\def\confName{CVPR}
\def\confYear{2022}

\begin{document}

\title{Context-Aware Sequence Alignment using 4D Skeletal Augmentation}

\author{Taein Kwon$^1$~~~~~~~~~~~~~~~~~~~
Bugra Tekin$^2$~~~~~~~~~~~~~~~~~~~
Siyu Tang$^1$~~~~~~~~~~~~~~~~~~~
Marc Pollefeys$^{1,2}$
\smallskip 
\\
$^1$Department of Computer Science, ETH Z\"urich~~~~~~$^2$Microsoft MR \& AI Lab, Z\"urich
}

\let\oldtwocolumn\twocolumn
\renewcommand\twocolumn[1][]{%
    \oldtwocolumn[{#1}{
    
    \begin{center}
           \includegraphics[width=0.95\textwidth]{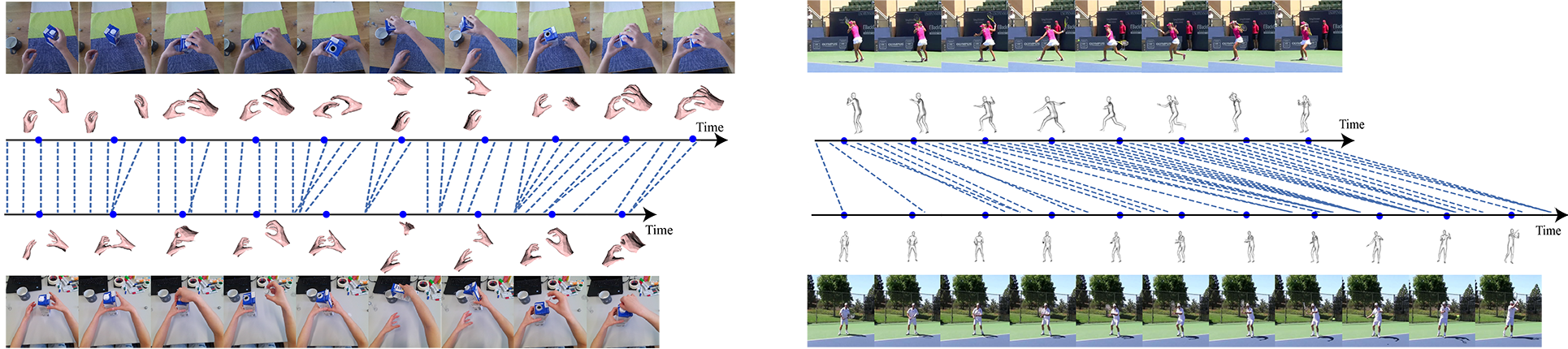}
           \vspace{-3mm}
           \captionof{figure}{{\bf Sequence Alignment.} %
           We propose a skeletal self-supervised learning approach that uses alignment as a pretext task.  
           Our work can align pose sequences for both hands and bodies, for which examples are shown above on the H2O~\cite{Kwon_2021_H2O} and PennAction~\cite{Zhang_2013_Penn} datasets. Our approach to alignment relies on a context-aware attention model that incorporates spatial and temporal context within and across sequences. Pose data provides a valuable cue for alignment and downstream tasks, such as phase classification and phase progression, as it is robust to different camera angles and changes in the background, while being efficient for real-time processing.
           } %
           
           \label{fig:teaser}
        \end{center}
    }]
}

\maketitle

\begin{abstract}
Temporal alignment of fine-grained human actions in videos is important for numerous applications in computer vision, robotics, and mixed reality.
State-of-the-art methods directly learn image-based embedding space by leveraging powerful deep convolutional neural networks. 
While being straightforward, their results are far from satisfactory, the aligned videos exhibit severe temporal discontinuity without additional post-processing steps.
The recent advancements in human body and hand pose estimation in the wild promise new ways of addressing the task of human action alignment in videos. 
In this work, based on off-the-shelf human pose estimators, we propose a novel context-aware self-supervised learning architecture to align sequences of actions. 
We name it {\it CASA}.
Specifically, CASA employs self-attention and cross-attention mechanisms to incorporate the spatial and temporal context of human actions, which can solve the temporal discontinuity problem.
Moreover, we introduce a self-supervised learning scheme that is empowered by novel 4D augmentation techniques for 3D skeleton representations.
We systematically evaluate the key components of our method. 
Our experiments on three public datasets demonstrate CASA significantly improves phase progress and Kendall's Tau scores over the previous state-of-the-art methods. 
\end{abstract}
\blfootnote{Project page: \href{https://www.taeinkwon.com/projects/casa/}{https://www.taeinkwon.com/projects/casa/}}

\input{01_introduction}

\input{02_related_work}

\input{03_method}

\input{04_evaluation}

\input{05_conclusion}

{\small
\bibliographystyle{ieee_fullname}
\bibliography{00_main}
}

\end{document}

\title{Supplementary Material: \\Context-Aware Sequence Alignment using 4D Skeletal Augmentation}

\author{Taein Kwon$^1$~~~~~~~~~~~~~~~~~~~
Bugra Tekin$^2$~~~~~~~~~~~~~~~~~~~
Siyu Tang$^1$~~~~~~~~~~~~~~~~~~~
Marc Pollefeys$^{1,2}$
\smallskip 
\\
$^1$Department of Computer Science, ETH Z\"urich~~~~~~$^2$Microsoft MR \& AI Lab, Z\"urich
}
\maketitle

\beginsupplement

In the supplemental material, we first provide details about the temporally smoothed noise we used for 4D augmentation and our hyper-parameters. Next, we analyze the performance of our approach for fine-grained frame retrieval and online sequence alignment. We then provide additional qualitative results of our algorithm for aligning two sequences and discuss our design choices for VPoser latent space, phase classification and encoding contextual informaton. Finally, we discuss the broader social impact  of our work. Additional qualitative visual results can be found in the video on the project page~\footnote{Project Page: \url{https://www.taeinkwon.com/projects/casa}}. Note that, in the video, we align the sequences by finding nearest neighbors in the embedding space without any post-processing. 

\subsection{Temporally Smoothed Noise}  
\begin{figure}[h]
\begin{center}
   \includegraphics[width=0.8\linewidth]{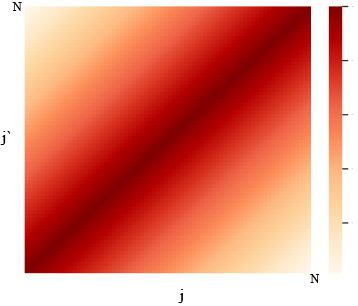}
\end{center}
\vspace{-6mm}
   \caption{\small {\bf Covariance matrix for our zero-mean multivariate normal noise distribution.}} 
 
\label{fig:cov_mat}
\end{figure}

While augmenting the joint angle and the latent space, the amount of noise applied across consecutive frames should not be completely independent of each other to preserve the smoothness and consistency of motion. To this end, we propose to add temporally smoothed noise, $MN(C)$, across the sequence, as explained in our paper. We model this using a multivariate normal distribution with a covariance matrix $C$ that enforces high correlation between temporally close frames within the same augmented sequence and low correlation between frames that are further away from each other. We depict the covariance matrix in Fig.~\ref{fig:cov_mat} and formulate it as follows:
\begin{equation}\label{eq:noise}
 C_{j,j'} = 1- \frac{\abs{ j-j'}}{2\cdot N},
\end{equation}
where $j$ and $j'$ depict two frame indices from the augmented sequence, and $N$ is the length of the augmented sequence. When $j$ and $j'$ are close to each other, the covariance is high, indicating that the noise applied on the poses at those frames are similar. This eventually results in less jittery and smooth augmented sequences.

\subsection{Implementation Details}
We list the hyperparameters we use in our experiments in Table~\ref{table:hyperparameters}.
\FloatBarrier
\begin{table}[H]
\begin{center}
\begin{adjustbox}{width=1.0\columnwidth,center}
\begin{tabular}{|l|r|}
\hline
Hyperparameter  & Value \\
\hline
Batch Size & 64 (Penn), 32 (H2O), 4 (IKEA)  \\
Learning rate & 3e-3(Penn), 3e-4 (H2O), 3e-2 (IKEA) \\
Optimizer & ADAM\\
Temperature ($\lambda_{temp}$) & 0.1\\
3D geometric noise probability & 0.3 \\
Noise standard deviation ($\sigma$) & 10 (angle), 0.1 (VPoser, translation)\\
Number of attention layers ($N_{att}$)& 4\\
Number of heads (parallel attention layers)&  15 (Penn), 17(IKEA), 21 (H2O)\\
Frames per second & 20 (Penn), 30 (IKEA, H2O) \\
\hline
\end{tabular}
\end{adjustbox}
\end{center}
\caption{\small {\bf Hyperparameters in our experiment.}}
\label{table:hyperparameters}
\end{table}
\FloatBarrier

\subsection{Fine-Grained Frame Retrieval}
\begin{table}
\begin{center}
\begin{adjustbox}{width=0.8\columnwidth,center}
\begin{tabular}{|c|c|c|c|}
\hline
Method  & AP@5 & AP@10 & AP@15 \\
\hline
SAL~\cite{misra2016shuffle} & 76.04&75.77&75.61  \\
TCN~\cite{sermanet2018time} &77.84&77.51&77.28  \\
TCC~\cite{dwibedi2019temporal} & 76.74&76.27&75.88  \\
LAV~\cite{haresh2021learning} &79.13&78.98&78.90  \\
CASA (Ours)& \bf{89.90} & \bf{89.44} & \bf{89.07} \\
\hline
\end{tabular}
\end{adjustbox}
\end{center}
\caption{\small {\bf Fine-grained frame retrieval.} We compare fine-grained frame retrieval results on the Penn Action dataset~\cite{Zhang_2013_Penn}.} %
\label{table:fine_retriev}
\end{table}

\begin{figure*}[]
\begin{center}
\includegraphics[width=0.8\textwidth]{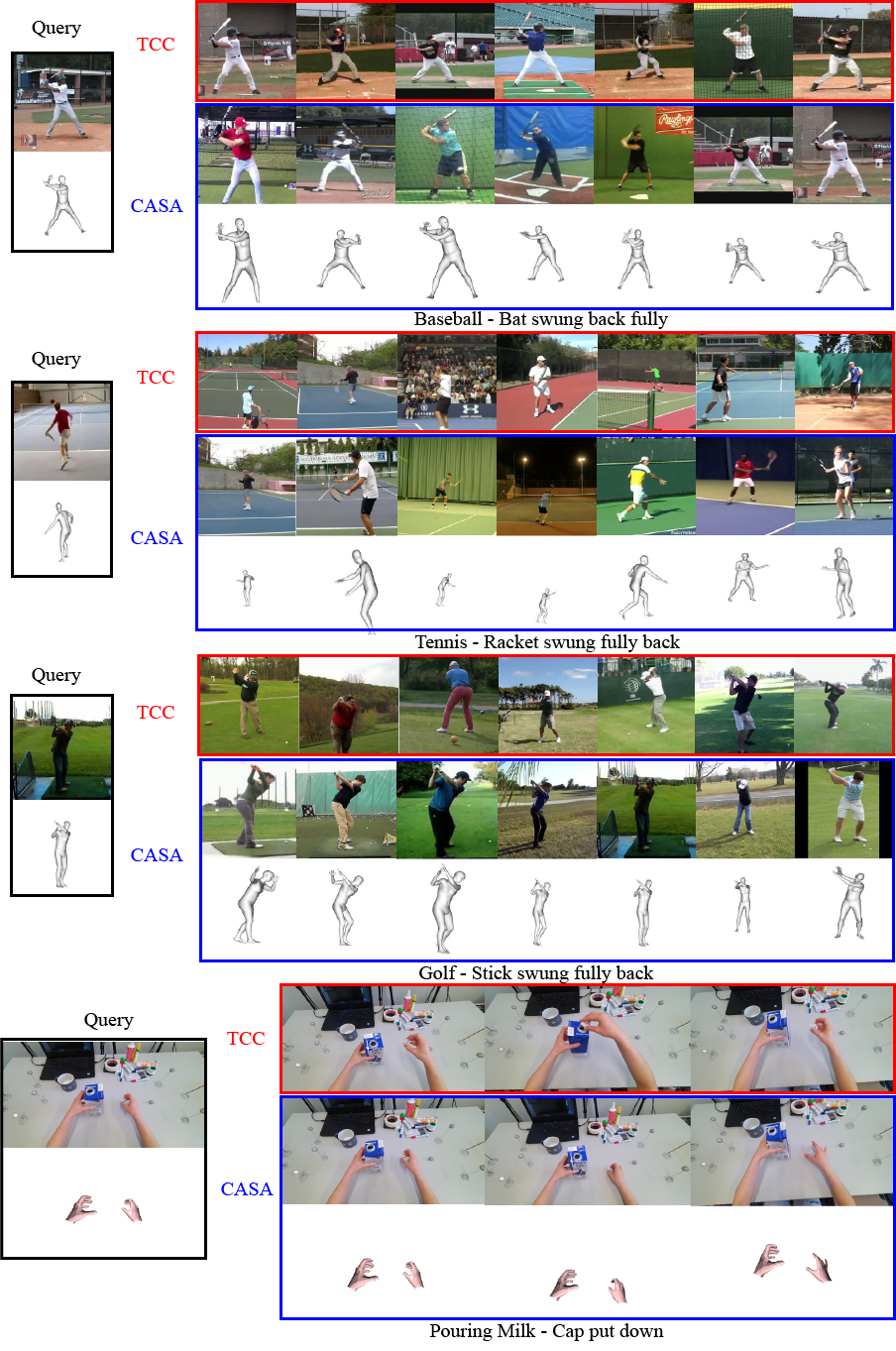}
\end{center}
\caption{\small {\bf Retrieval results.} We visualize our fine-grained retrieval results for the Penn Action ($k=7$) and H2O (k=3) datasets in comparison to TCC~\cite{dwibedi2019temporal} and demonstrate that our method is able to successfully retrieve visually similar frames. For example, in the ``tennis'' sequences, our method is able to find ``racket swung fully back" action correctly in all the examples whereas TCC fails to retrieve it in some of them.}
\label{fig:retrieval}
\end{figure*}

We show fine-grained frame retrieval results in Table~\ref{table:fine_retriev}. %
We find the K nearest frames from one query frame in the embedding space. Following ~\cite{haresh2021learning}, we report Average Precision (AP) at K, that is, the average percentage of correctly retrieved action phase labels within K-retrieved frames. Table~\ref{table:fine_retriev} shows that our method improves upon prior work by a large margin (an improvement of $10.77\%$ at $K=5$, $10.46\%$ at $K=10$ and $10.17\%$ at $K=15$).
In Fig.~\ref{fig:retrieval}, we show qualitative results of our algorithm compared to TCC~\cite{dwibedi2019temporal}. We observe that our method is able to accurately retrieve relevant frames by reasoning about the temporal context of the actions.

\subsection{Online Sequence Alignment}
Our method uses an attention-based model to capture context from all the frames to compute alignment across two videos. However, for online applications, the assumption of having the full sequence will not be valid. Therefore, to demonstrate the potential of our approach for online applications (\eg, online task guidance in augmented reality), we perform an additional experiment, in which, we use contextual information only using frames, seen until the current time frame. To this end, we rely on embeddings computed until the current frame and use it for matching across sequences. Table~\ref{table:online} demonstrates that we report consistently high sequence alignment performance as compared to existing approaches, even when we perform at a fully online manner only using contextual information from past frames.

\begin{table}
\begin{center}
\begin{adjustbox}{width=0.8\columnwidth,center}
\begin{tabular}{|c|c|c|c|c|}
\hline
\shortstack{} & Offline  & Online    & TCC~\cite{dwibedi2019temporal} & LAV~\cite{haresh2021learning} \\
\hline
Phase classification & 92.20& 88.01 &81.35 &84.25\\
Phase progress& 0.9449& 0.8454   &0.6638 & 0.6613\\
Kendall's Tau ($\tau$) & 0.9728& 0.9059  &0.7012 &0.8047\\
\hline
\end{tabular}
\end{adjustbox}
\end{center}
\vspace{-3mm}
\caption{\small {\bf Ablation study of online sequence alignment.}
We compare the phase classification, phase progress, and Kendall's tau for online and offline operating modes of our model on the Penn Action dataset~\cite{Zhang_2013_Penn}. %
While we use the full sequence for offline mode, we only use the embeddings up until the current frame for the online mode.}
\label{table:online}
\end{table}

\subsection{Qualitative Results}
\begin{figure*}[]
\begin{center}
\includegraphics[width=1.0\textwidth]{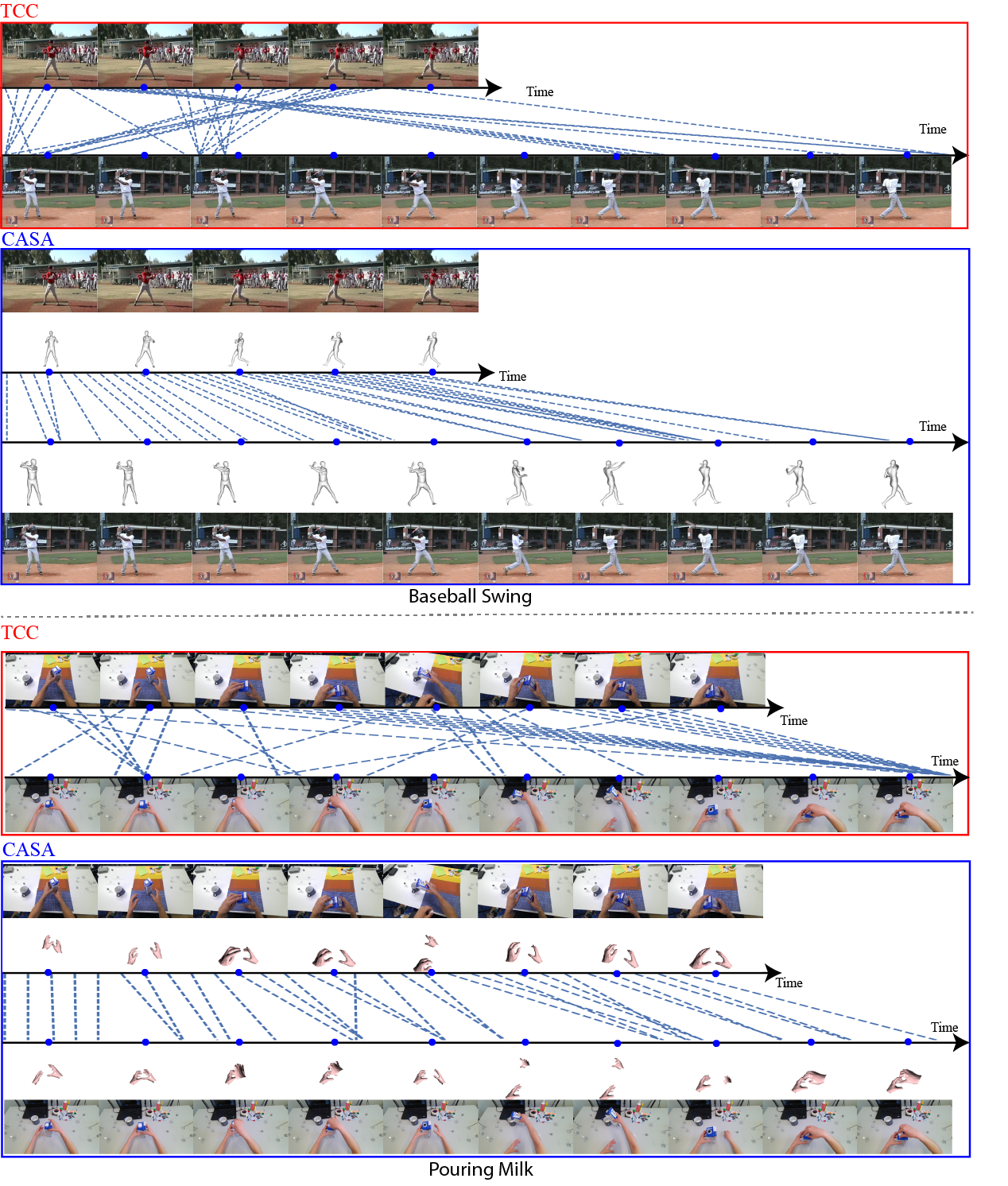}
\end{center}
\caption{\small {\bf Sequence alignment results}. We draw matching lines for every 20 frames in the \emph{pouring\_milk} sequence and for every frame in \emph{baseball\_swing} sequence.}
\label{fig:alignment}
\end{figure*}

\begin{figure*}[]
\begin{center}
\includegraphics[width=1.0\textwidth]{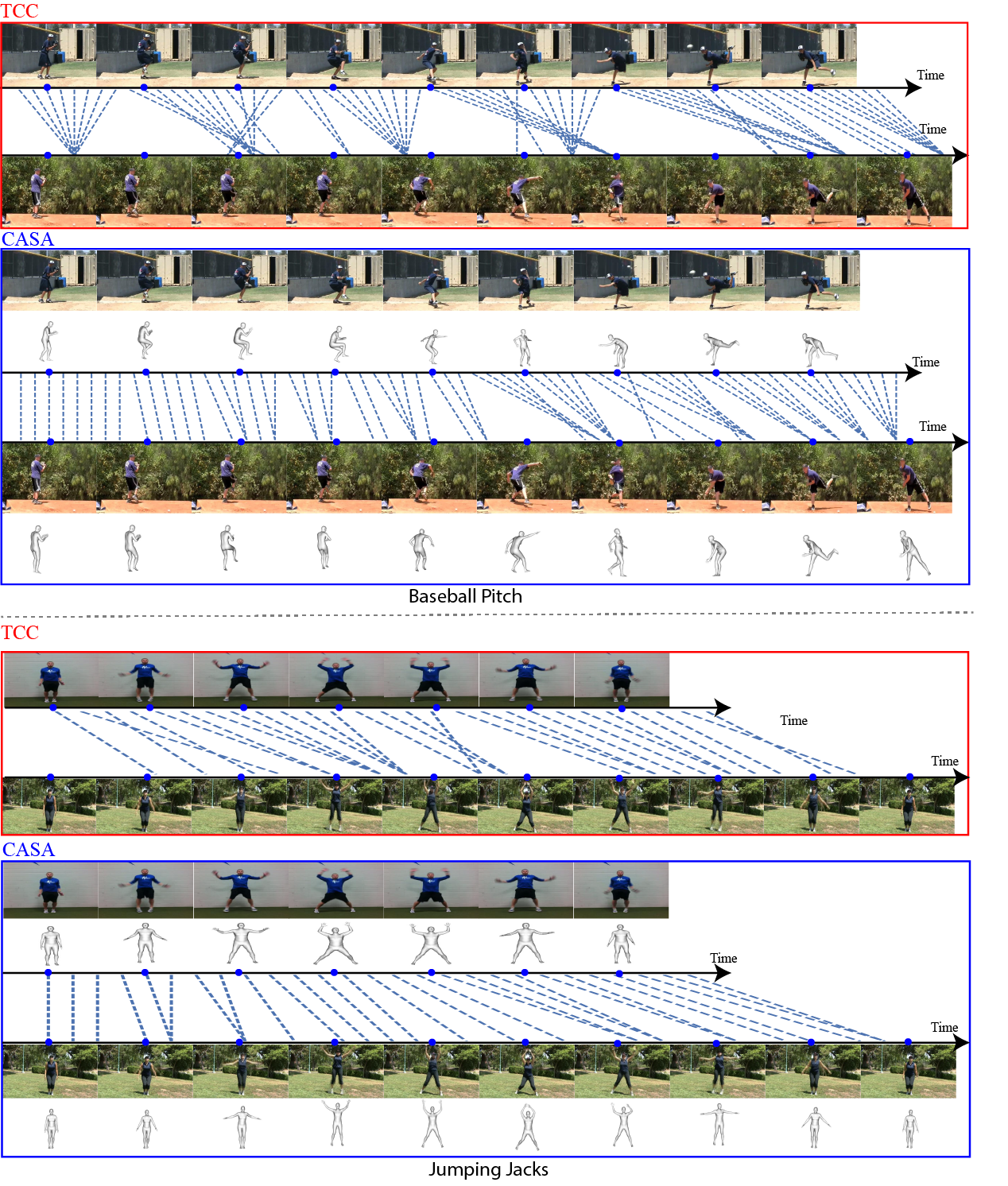}
\end{center}
\caption{\small {\bf Sequence alignment results}. We draw matching lines for every frame in the \emph{baseball\_pitching} and \emph{jumping\_jacks} sequences.}
\label{fig:alignment2}
\end{figure*}

We provide additional sequence alignment results of our approach in Fig.~\ref{fig:alignment} and Fig.~\ref{fig:alignment2}. Our method is able to align sequences across time by considering temporal context.

\begin{figure}[t]
\begin{center}
   \includegraphics[width=1.0\linewidth]{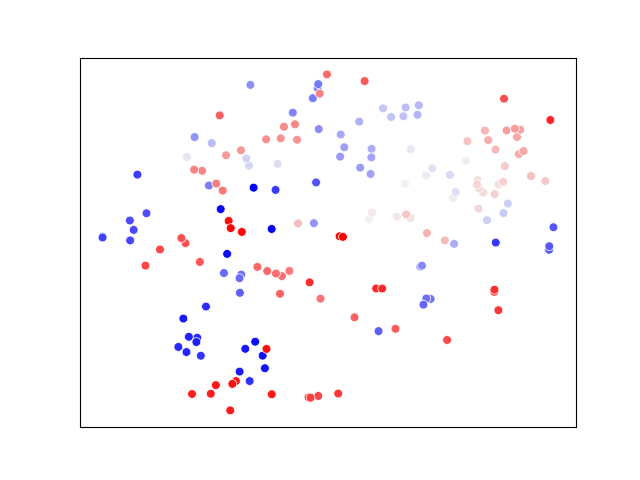}
\end{center}
\vspace{-6mm}
   \caption{\small {\bf t-SNE distribution.} We visualize the t-SNE distribution using \emph{baseball\_pitch} action on the PennAction dataset. Each point represents an encoded pose using VPoser~\cite{Pavlakos2019SMPL-X}. We use the same color for the poses in the same sequence. In addition, we show the beginning and end frames with different shades of the same color (\eg first frames are encoded with lighter colors, while later frames are encoded with darker colors.)} 
 
\label{fig:tsne_distribution}
\end{figure}

\subsection{VPoser Latent Space}
 In Fig.~\ref{fig:tsne_distribution}, we observe from the t-SNE visualization of the pose embedding of VPoser~\cite{Pavlakos2019SMPL-X} that the latent space is smooth and well-behaved. Different sequences with the same action are embedded in closeby locations in the embedding space and temporally close frames are mapped to nearby points in the
embedding space. Therefore we conjecture that augmentations applied on the VPoser latent space would correspond to reasonable spatiotemporal augmentations in the original pose space and would enrich our dataset with diverse and realistic pose sequences.
 
 \subsection{SVM vs Nearest Neighbor}
 
 In our main paper, we provide results for phase classification on features learned through self-supervised learning using SVM classifier. We further compute phase classification results using nearest neighbor which does not require any training data. We obtain an accuracy of 89.52\% on the PennAction dataset, which is still beyond the state-of-the art, even without using any training data. Note that the previous state-of-the-art method (LAV)~\cite{haresh2021learning} reports 83.56\%, 83.95\% and 84.25\% phase classification accuracy, when using an SVM classifier trained on a fraction of 10\%, 50\% and 100\% of the ground
truth labels.
 
 \subsection{Encoding Contextual Information}
 
The global receptive field, positional encoding and self- and cross-attention layers of Transformer enable the transformed feature representations to be context- and position-dependent, as also posited by prior work~\cite{carion2020end,sun2021loftr}. Therefore our architecture enables our method to be aware of the spatial and temporal context of the human actions.

\subsection{Societal Impact}

While our method for sequence alignment provides many beneficial use cases for AR-based task guidance, it could also be misused for surveillance and monitoring people's actions. This could raise privacy concerns and therefore use of this technology should be guided by responsible AI principles.

{\small
\bibliographystyle{ieee_fullname}
\bibliography{06_supp}
}

%% file: 01_introduction.tex
\section{Introduction}\label{sec:introduction}

Temporal alignment of human activities in videos aims to identify sequential per-frame correspondence between two video instances of the same action as shown in Fig.~\ref{fig:teaser}. 
This is challenging due to large variation in speed of actions, severe self-occlusion, and diverse backgrounds across different videos. 
Furthermore, an accurate temporal alignment of human activities requires semantic understanding of human motion and causal reasoning of the action stages.
When it comes to  hand-centric fine-grained activities under first-person views, the challenges are amplified by the varying viewpoints and embodied movement of camera wearers.
State-of-the-art methods leverage large-scale datasets and powerful deep convolution neural networks to learn image-based representation to perform temporal video alignment \cite{dwibedi2019temporal, haresh2021learning}. Despite rapid progress in terms of accuracy and advanced learning schemes, the results %
are still far from applicable to real-world applications.

Recent advancements and growing availability of head-mounted devices (e.g.~Microsoft HoloLens~\cite{ungureanu2020hololens}) enable new ways of communication and collaboration. For instance, the built-in hand tracking system of HoloLens provides real-time accurate hand pose estimation of the camera wearer. Such systems promise a revolution in how hand motion and actions can be captured, modeled, and analyzed. Consequently, they point towards a new way to align fine-grained hand-centric actions in videos based on 3D skeleton motion extracted from  off-the-shelf pose estimators. 

One appealing application of this setting is to utilize mixed reality headsets to close the skills gap between experts and learners. 
Traditionally, transferring skills from experts to learners is not easy. 
Experts often have to stay close to learners to teach and inspect individuals.
Given the videos shared by experts from their point-of-view and on-device hand pose estimation, an accurate temporal alignment method that provides dense correspondences between the fine-grained hand actions performed by the expert and the learners will enable significantly more efficient and precise skill transfer guidance.

\begin{figure}[]
\begin{center}
   \includegraphics[width=1.0\linewidth]{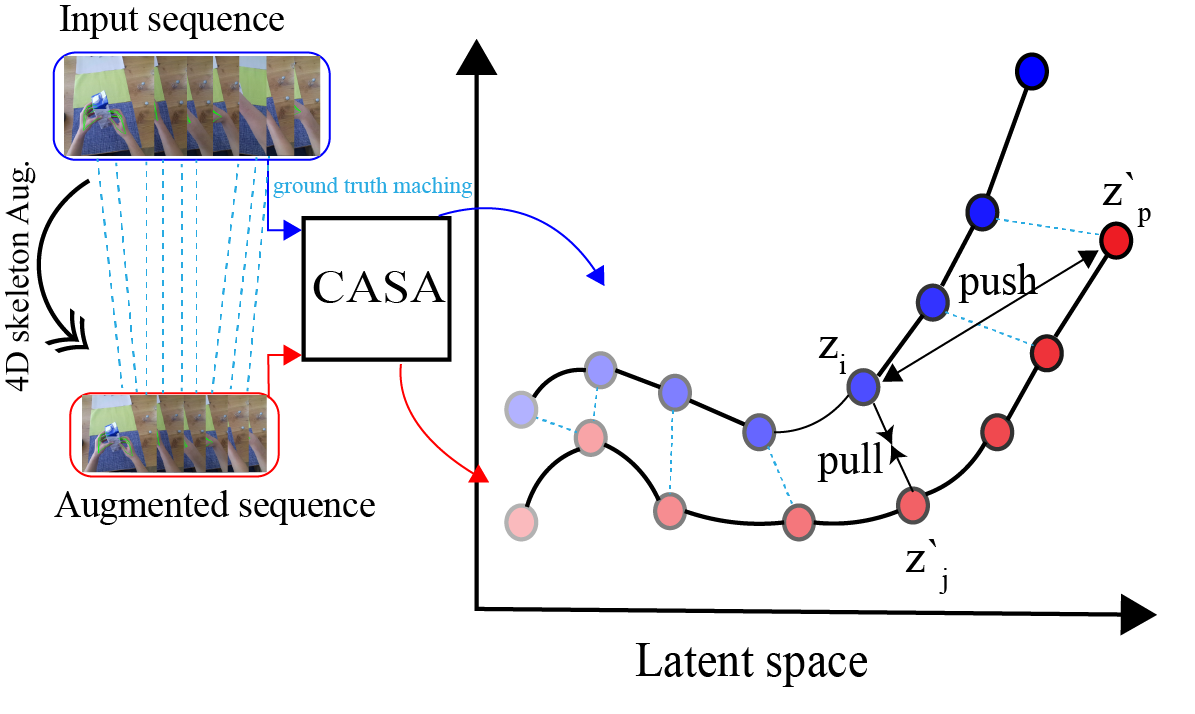}
\end{center}
\vspace{-6mm}
   \caption{\small {\bf Self-supervised learning using 4D augmentation.} Given a sequence and its augmentation in 4D, we optimize our latent space such that the distances between the features of matching frames ($z_{i}$ and $z'_{j}$ ) are minimized, while those of non-matching frames  ($z_{i}$ and $z'_{p}$ ) are encouraged to be further apart. 
   }
   \vspace{-4mm}
   
\label{fig:keyidea}
\end{figure}

Inspired by these observations, we propose to align 3D skeletons extracted from videos for human action alignment tasks. Instead of using 2D features used in \cite{dwibedi2019temporal,haresh2021learning,purushwalkam2020aligning}, %
we propose CASA, Context-Aware Sequence Alignment, a novel context-aware self-supervised learning framework for 3D skeletons using 4D augmentation. 
As shown in Fig.~\ref{fig:keyidea}, our framework reasons about the context through the attention module and performs self-supervised learning with our novel 4D augmentation strategies. %
From the ground-truth matching between the augmented and original sequences, we can learn powerful representation to perform downstream tasks. 

Furthermore, our 3D skeleton-based alignment method not only works for hand action analysis but also can be applied to full body-related actions where we extract 3D human bodies from videos using off-the-shelf body estimators. Although, in such cases the reconstructed 3D human bodies can be less accurate than the hand tracking result from mixed reality devices, our method still generalizes well due to the novel context-aware network architecture and the self-supervised learning framework enabled by the powerful 4D augmentation schemes.

We perform extensive experiments to validate the effectiveness and applicability of CASA on three public datasets: Penn Action~\cite{Zhang_2013_Penn}, IKEA ASM~\cite{Ben-shabat_2020_Ikea}, and H2O~\cite{Kwon_2021_H2O}. CASA achieves  the  best  performance  in most phase classification tasks of three datasets. 
Furthermore, in terms of phase progress and  Kendall's tau, our method significantly outperforms the previous state-of-the-art methods~\cite{dwibedi2019temporal,haresh2021learning}. 
The results demonstrate the importance of knowing the context of action and the applicability of utilizing 3D poses for fine-grained video alignment tasks.

\boldparagraph{Contributions.}
In summary, our contributions are:
\textbf{(1)} we propose a novel attention-based and context-aware dense alignment framework for fine-grained human action analysis;
\textbf{(2)} we introduce novel 4D augmentation strategies for 3D skeletons in self-supervised learning that consider both temporal and spatial augmentation; %
\textbf{(3)} to the best of our knowledge, it is the first work to perform 3D skeleton-based fine-grained video alignment using self-supervised learning. We prove the utility of our 3D skeleton-based temporal alignment methods by largely outperforming the state-of-the-art in three public datasets. %

%% file: 02_related_work.tex
\section{Related Work}\label{sec:related_work}

\boldparagraph{Self-Supervised Learning.} Several image-based self-supervised learning methods have been proposed recently that rely on different hand-crafted pretext tasks. For example, recent work used image colorization~\cite{larsson2016learning}, solving jigsaw puzzles ~\cite{noroozi2016unsupervised,wei2019iterative}, rotation prediction~\cite{gidaris2018unsupervised} or image inpainting~\cite{jenni2020steering} as pretext tasks to train self-supervised models. These handcrafted tasks rely on particular adhoc heuristics, which limits their generalization power. Alternatively, contrastive learning approaches learn representations by contrasting positive pairs against negative pairs~\cite{dosovitskiy2014discriminative,wu2018unsupervised,zhuang2019local,tian2020contrastive,he2020momentum,misra2020self}. Notably, Chen et al.~\cite{chen2020simple} demonstrated that composition of multiple data augmentation operations is crucial in defining the contrastive prediction tasks that yield effective representations for single image data. Inspired by the success of self-supervised methods in image domain, recently several self-supervised learning methods were proposed for videos, either using pretext tasks, such as predicting future frames~\cite{ahsan2018discrimnet,diba2019,srivastava2015unsupervised,vondrick2016generating}, clip order~\cite{fernando2017self,lee2017unsupervised,misra2016shuffle,xu2019self}, pace~\cite{benaim2020speednet,cho2020self,wang2020self,yao2020video} or arrow of time~\cite{pickup2014seeing,wei2018learning}, or focusing on instance-based contrastive learning techniques~\cite{dave2021tclr,hu2021contrast,feichtenhofer2021large,qian2021spatiotemporal}.

Compared to image and video-based self-supervised learning, skeleton-based self-supervised learning started to emerge as an active field only recently. Proxy tasks such as skeleton inpainting~\cite{zheng2018unsupervised} and motion prediction~\cite{su2020predict} have been proposed by recent work. However, such methods do not explicitly account for the spatio-temporal dependencies of skeletal representations.  
Skeletal self-supervised learning techniques that rely on neighborhood consistency~\cite{si2020adversarial}, fusion of multiple pretext tasks~\cite{lin2020ms2l} and motion continuity~\cite{su2021self} have also shown the promise of self-supervised techniques for learning skeletal sequence representations.  Unlike previous approaches, we propose a self-supervised learning framework with a composition of 4D data augmentation strategies. We consider both temporal and spatial transformations of the data, globally for the skeletal motion, and locally for individual joints.

\boldparagraph{Transformer.}
After the success of the Transformer architecture~\cite{vaswani2017attention} in Natural Language Processing (NLP), there has been a surge in interest in its application for computer vision. Several Transformer-based architectures have been proposed for image classification~\cite{dosovitskiy2020image}, object detection~\cite{carion2020end}, and semantic segmentation~\cite{wang2020axial}. More closely related to our work, Sun et al. and Sarlin et al. \cite{sun2021loftr,sarlin2020superglue} proposed Transformers for the task of image alignment. While Transformers have been actively used within supervised learning contexts, recent work also has shown the potential of self-supervised pretraining of a standard Vision Transformer model for several downstream tasks~\cite{caron2021emerging}.  Correspondingly, in this work, we propose a self-supervised Transformer architecture for fine-grained alignment of videos.

\boldparagraph{Sequence Alignment.}
Dynamic Time Warping (DTW) has become the \emph{de facto} standard for unsupervised sequence matching due to its simplicity and generality for different types of modalities~\cite{berndt1994using}. Cuturi and Blondel~\cite{cuturi2017soft} proposed a differentiable approximation of DTW which allows for pairing it with neural networks and training sequence models. Canonical Time Warping~\cite{zhou2009canonical} and Generalized Time Warping~\cite{zhou2012generalized} generalized DTW and enabled alignment of signals with different dimensionality. As an alternative to DTW, Su et al.~\cite{su2017order}, relied on optimal transport to match two sequences frame-by-frame, while regularizing the loss such that temporal information is preserved in the matching process. While focusing on the alignment problem, these approaches do not aim at feature learning for sequence matching unlike our work.

Closely related to the sequence alignment problem, metrics for assessing human motion similarity have been actively explored by previous studies~\cite{berndt1994using,dong2020motion,liu2021normalized,martinez2017human,coskun2018human,sun2020view,sumer2017self,mori2015pose,sun2020view}. The assessment of the similarity between two sequences of poses or motion is a non-trivial problem since human motion varies across  sequences due to a number of different factors such as speed, anthropometric variations, and subject-specific pose patterns. Conventional approaches for measuring similarity of human motion sequences are based on estimating the L2 displacement error~\cite{martinez2017human,dong2020motion} or DTW~\cite{berndt1994using}. However, these metrics disregard  contextual information in the time dimension, which limits their application for human motion analysis. To overcome the limitations of standard metrics, deep metric learning methods have been proposed by~\cite{coskun2018human,sun2020view,sumer2017self,mori2015pose}.

In the context of self-supervised learning-based video alignment~\cite{sermanet2018time,dwibedi2019temporal,haresh2021learning,hadji2021representation}, Time Contrastive Networks (TCN)~\cite{sermanet2018time} used synchronized frames with contrastive learning to align frames from different points of view. Temporal Cycle Consistency (TCC) method~\cite{dwibedi2019temporal} learned an embedding space that maximizes one-to-one mapping of cycle-consistent points across pairs of video sequences. Learning by Aligning Videos (LAV)~\cite{haresh2021learning} adopted soft-DTW~\cite{cuturi2017soft} as a self-supervised temporal alignment loss. Unlike our work that aims at self-supervised skeletal sequence learning, these works all focused on matching images across videos.

%% file: 03_method.tex
\begin{figure*}[]
\begin{center}
   \includegraphics[width=1.0\linewidth]{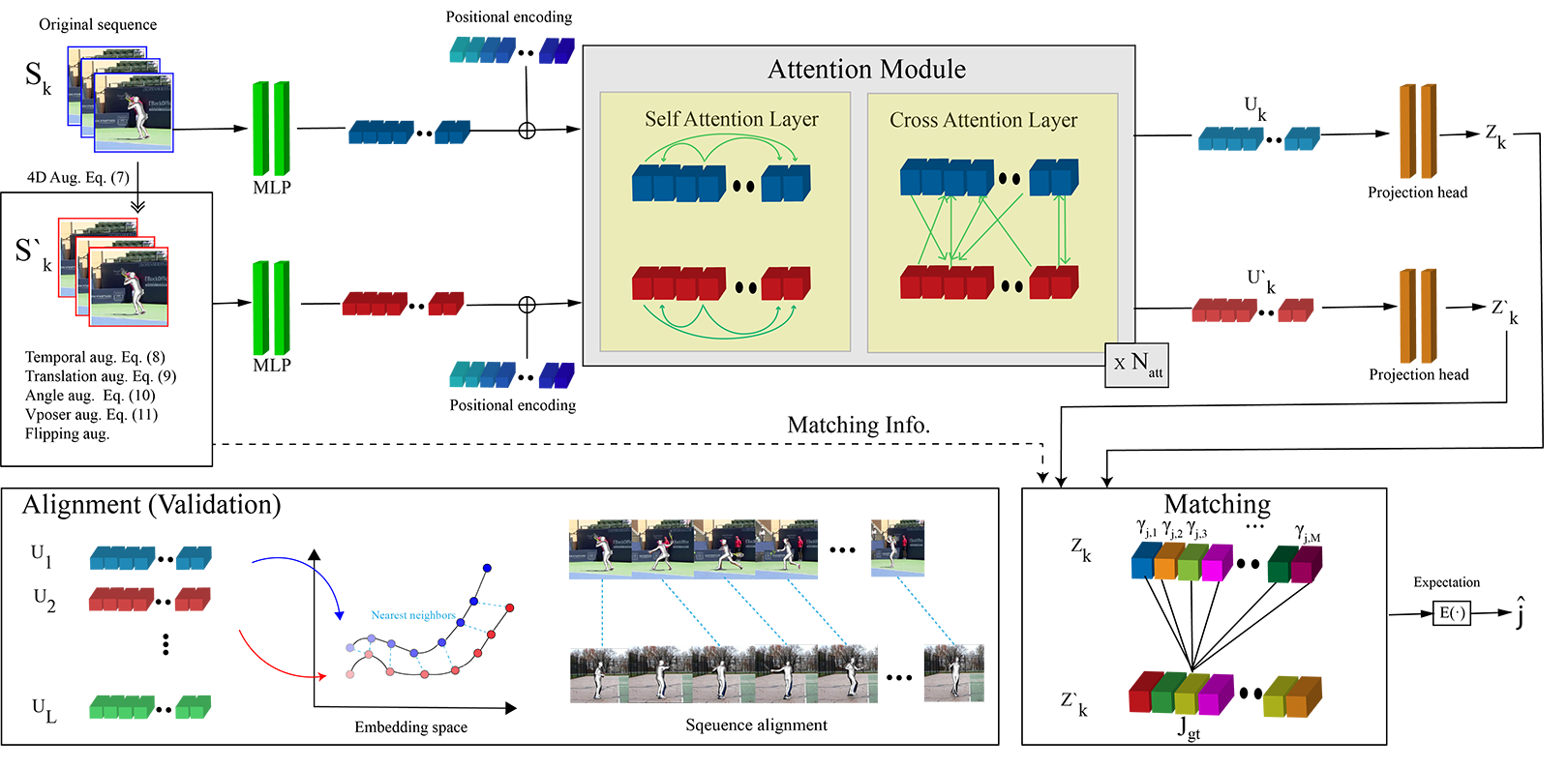}
\end{center}
\vspace{-6mm}
   \caption{\small {\bf Overview of our pipeline.} The proposed framework takes as input a skeleton sequence $S_k$ along with its spatio-temporally augmented version $S'_k$. Both sequences are encoded by temporal positional encodings. 
   Self- and cross-attentional layers learn contextual information within and across sequences with the help of temporal positional encoding.
   We employ a projection head to improve our representation quality~\cite{chen2020simple} . We use a \emph{contrastive regression loss} that matches a pose sequence with its 4D augmented version. For the downstream tasks and alignment, we use the embeddings before the projection head stage.}
\vspace{-4mm}
\label{fig:pipeline}
\end{figure*}
\section{Method}\label{sec:method}

Fig.~\ref{fig:pipeline} shows an overview of our proposed pipeline. We propose a self-supervised skeletal representation learning approach that uses skeletal alignment as a pretext task. Our model relies on an attention-based context-aware framework for sequence alignment. Our self-supervised loss, inspired by the success of image-based contrastive learning~\cite{chen2020simple}, relies on minimizing the difference between a skeletal sequence and its augmentation in 4D, that is, in 3D space and time. Our framework learns a representative latent space, which is effective in downstream tasks and can be used to align two skeletal sequences via nearest-neighbor search.

\boldparagraph{Notations.} 
Each 3D skeleton of a sequence is defined as $s_i \in \mathbb{R}^{J \times 3}$ with J skeleton joints in $x,y,z$ locations. 
Each $k$-th ($1 \leq k \leq L$) sequence of skeletons is shown with $S_k = \{s_1,s_2,..,s_M\}$ and its augmentation is shown with $S_k' = \{s'_1,s'_2,..,s'_N\}$. 
The embedding of a skeletal sequence is computed as $(U_k,U'_k)=\Phi(S_k,S'_k;\Omega)$, where $\Phi$ is our framework's encoder network with the parameters, $\Omega$.
The embedding of the original sequence, $U_k$, is denoted with $\{u_1,u_2,...u_M\}$  and that of the augmented sequence, $U'_k$, is denoted with $\{u'_1,u'_2,...u'_N\}$. 
Our latent space in which we optimize an alignment loss is denoted with ($Z_k=\{z_1,z_2,...z_M\}=P(U_k)$ and $Z'_k=\{z'_1,z'_2,...z'_N\}=P(U'_k)$), where $P(\cdot)$ is a projection head~\cite{chen2020simple}. 
Note that we use upper-case notations for sequence-level processing and lower-case notations for per-frame processing.

\subsection{Preliminaries}

\boldparagraph{3D human body representation.}
We use SMPL~\cite{Pavlakos2019SMPL-X} on the Penn Action dataset and Keypoint RCNN~\cite{he2017mask} body joints representation,~$s_{rcnn}, \in \mathbb{R}^{17\times3}$ on the IKEA dataset. Pose parameters,~$\theta_{smpl} \in \mathbb{R}^{72}$, store angles for 22 skeleton joints along with a global rotation and translation vector.
We remap the 22 SMPL skeleton joints to the skeleton representation of FrankMocap~\cite{rong2021frankmocap},~$s_{smpl} \in \mathbb{R}^{25\times3}$, to be able to  use the FrankMocap estimator. We recover 3D body skeleton,~$s_{smpl}$, based on  $SMPL(\beta_{smpl},\theta_{smpl})$, where $SMPL(\cdot)$ is the function that calculates 3D skeleton, given shape, $\beta_{smpl}$, and pose, $\theta_{smpl}$, parameters.  

\boldparagraph{3D hand representation.}
We use MANO~\cite{Romero2017MANO} 3D hand skeleton representation~$s_{mano} \in \mathbb{R}^{42\times3}$ in the H2O dataset. 
MANO contains human hand shape parameters,~$\beta_{mano} \in \mathbb{R}^{20}$, and pose parameters,~$\theta_{mano} \in \mathbb{R}^{34\times3}$, storing angles for 30 skeleton joints, 2 global rotation and 2 translation vectors for both hands.  We recover 3d hand skeleton~$s_{mano}$ from $MANO(\beta_{mano},\theta_{mano})$, where $MANO(\cdot)$ is a function that calculates 3D hand skeletons given shape ($\beta_{mano}$ ) and pose parameters ($\theta_{mano}$).

We will call, $s_{smpl}$, $s_{rcnn}$, and $s_{mano}$ as $s$ and omit $\beta$ from the following equations for simplicity. Also, we will use the transform function~$T(\cdot)$ instead of $MANO(\cdot)$ and $SMPL(\cdot)$. Accordingly, we will transform each pose parameter to 3D skeletons using $S_k=T(\Theta_k)$.

\subsection{Model architecture}

Our model consists of  multi-layer perceptrons (MLP), positional encodings, an attention module, and projection heads. In the following, we will explain each part of the model.

\boldparagraph{MLP.}
We use two nonlinear layers of a fully connected network with the same input dimension to extract features from our 3D joint representation before feeding them input to the attention module. 

\boldparagraph{Attention module.}
Transformer~\cite{vaswani2017attention} has received a lot of attention due to its impressive performance in the NLP field, as summarized in Section~\ref{sec:related_work}. To leverage the power of Transformers in temporal understanding, we employ self- and cross- attention layers, that efficiently capture temporal context as compared to methods that compute features from single-images~\cite{dwibedi2019temporal,haresh2021learning}. We model self-attention to learn dependencies within the skeletons in the same sequence and cross-attention to learn the inter-dependencies between the original sequences and their 4D augmentations. To reduce the computational complexity of attention layers, we adopt Linear Transformer~\cite{katharopoulos2020transformers} architecture. %
By contrast to the earlier Transformer-based works~\cite{sarlin2020superglue,sun2021loftr}, our attention module is integrated into a self-supervised learning framework that uses 4D augmentations for sequence matching.

\boldparagraph{Temporal positional encoding.} %
We inject temporal information into our framework using positional encodings~\cite{vaswani2017attention}. Using  positional encodings, our model reasons about temporal locations of each skeleton frame. Such information is crucial in understanding temporal dependencies between skeletons. Different from other vision-based tasks~\cite{dosovitskiy2020image, sun2021loftr}, we only need 1D positional encodings as the order of joints in the skeleton is fixed. We choose sinusoidal positional encoding as it is proved to be effective in machine translation, which can be conceptually similar to aligning two skeletal sequences from the same activity.  

\begin{equation}\label{eq:positional_encoding}
  PE_i = \begin{cases} sin(w_l \cdot i),  i=2l \\
                       cos(w_l \cdot i),  i=2l+1 
        \end{cases},
\end{equation}

where $w_l = \frac{1}{5000^{(2l/d)}}$ and $d$ is the dimension of the skeleton joints, $i$ is an index for the temporal frame location in the sequence. We choose $5000$ for the denominator of $w_l$ since the maximum length of the sequences in our case is bounded by 5000.

\boldparagraph{Projection head.} To improve the quality of our representation, we employ a projection head as in ~\cite{chen2020simple}. As shown by~\cite{chen2020simple}, without the projection head, the learned model is more likely to overfit to the optimization task. While we optimize for the alignment, we aim to have representative features for downstream tasks that address fine-grained action recognition. Therefore we use a projection head, $p(\cdot)$, in the form of an MLP with one hidden layer.

\begin{equation}\label{eq:projection_head}
  z_i = p(u_i) = W^2\sigma(W^1u_i), 
\end{equation}

where $\sigma$ is a ReLU layer and $W^1$ and $W^2$ are fully connected layers. We show that the projection head improves our accuracy in downstream tasks in Section~\ref{sec:ablation}.

\paragraph{Matching and loss.}\label{sec:loss}
Given an original sequence and its augmentation in the time dimension, the temporal correspondences between two sequences are already known and preserved. Note also that 3D geometric augmentation will not affect the correspondences between two sequences as the 3D perturbations we use for data augmentation are time-independent.  
Our self-supervised learning framework, inspired by recent advances in contrastive learning~\cite{chen2020simple, hadsell2006dimensionality}, learns representations by maximizing the agreement between positive pairs, which we take, in our case, as a skeletal sequence and its 4D augmentation.
We formulate the contrastive loss for positive pairs, $(i,j)$, using the following equation:
\begin{equation}\label{eq:contrastive_loss}
  \mathcal{L}_{i,j} = -log\frac{exp(-\norm{z_i-z_j}/\lambda_{temp})}{\sum_{m=1}^{N}exp(-\norm{z_i-z_m}/\lambda_{temp})},
\end{equation}
where $\lambda_{temp}$ is a temperature parameter.
However, the classification-based loss can not reason about how far the prediction for a matched frame is, from the ground-truth alignment. Therefore, instead of using Equation~\ref{eq:contrastive_loss}, we adopt a regression loss~\cite{dwibedi2019temporal} to penalize nearby frames less by accounting for the temporal relationships of neighboring frames. 
The difference from~\cite{dwibedi2019temporal} is that we compute this loss for every frame to gather contextual information from the whole sequence, instead of using frames only from a local neighborhood. 
The probability of a frame, $i$, in the original sequence, being a match to a frame $j$, in the augmented sequence, is denoted with $\gamma_{j,i}$ and computed by 
\begin{equation}\label{eq:softmax}
  \gamma_{j,i}=\frac{e^{-\norm{z'_j - z_i}/\lambda_{temp}}}{\sum_{m=1}^{M} e^{-\norm{z'_j - z_m}/\lambda_{temp}}},
\end{equation}
where $\gamma_{j,i}$ is $i$-th value of probability $\gamma_{j}$. We then predict the target frame index, $\hat{j}$, by weighing the frame indices with their corresponding probabilities, as follows:
\begin{equation}\label{eq:predict}
  \hat{j} = \sum_i^M (\gamma_{j,i}\cdot i),
\end{equation}
The final loss $\mathcal{L}$ will be the mean squared error between the predicted frame index $\hat{j}$ and ground truth frame index $j_{gt}$, which is already known and preserved after data augmentation.
\begin{equation}\label{eq:loss}
  \mathcal{L} = \frac{1}{N}\sum_j^N \norm{j_{gt}-\hat{j}}^2,
\end{equation}
\subsection{4D Augmentation}

\begin{figure}
\begin{center}
   \includegraphics[width=1.00\linewidth]{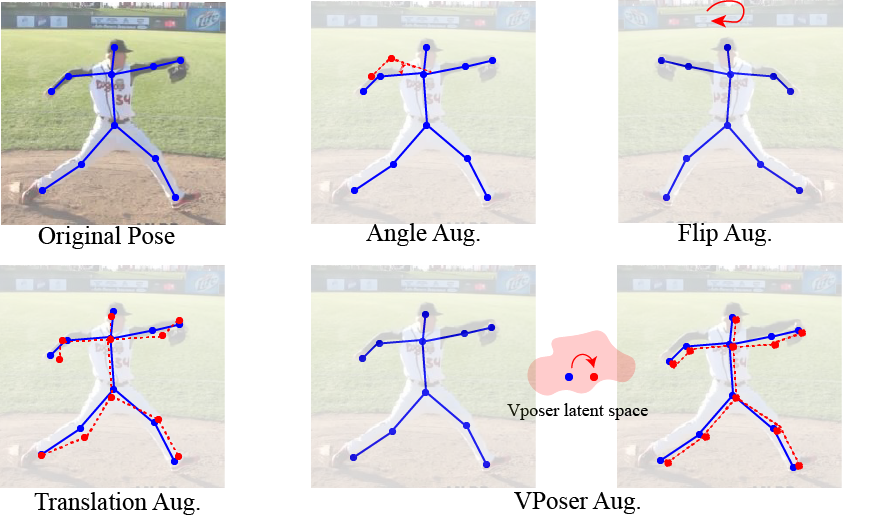}
\end{center}
\vspace{-4mm}
   \caption{\small {\bf Different types of 3D geometric augmentation.} Translation augmentation tackles the noisy estimates of off-the-shelf pose estimators. We observe that different augmentation strategies produce feasible poses that provide positive matches for the self-supervised learning framework.}
   \vspace{-4mm}
\label{fig:4d_aug}
\end{figure}

To be able to create positive pairs of skeletal sequences with known correspondences, we propose to augment the skeletal sequences in 3D space and time. 
We illustrate our proposed 4D augmentation strategies in Fig.~\ref{fig:4d_aug}.  We propose 5 different augmentation schemes: temporal augmentation, joint angle augmentation, translation augmentation, skeleton flipping, and augmenting the latent space of skeletons based on VPoser~\cite{Pavlakos2019SMPL-X}. 
We perform the augmentation by adding noise to each skeletal joint translation or angle. 
To be able to generate realistic augmentations of skeletons that would reflect different variations of motion, we propose to add temporally smoothed noise across the sequence, using a multivariate normal distribution, which has a covariance matrix that contains high correlations along the diagonal such that temporally closer points are highly correlated. We provide further details about the distribution we employ for temporally smoothed noise in our supplemental material.
We apply temporally smoothed noise for augmenting joint angles and the latent space obtained by VPoser. This strategy overall enables the motion to be continuous and smooth over time. The augmentation function $G(\cdot)$ is defined as:
\begin{equation}\label{eq:4D_augmentation}
 S'_k = G_{temp,trans,flip}(T(G_{VPoser,angle}(\Theta_k))), \\
\end{equation}
In what follows, we describe our different augmentation strategies in more detail.

\boldparagraph{Temporal augmentation.}
We randomly select N frames in the original M frames. Through this step, our self-supervised learning framework learns the different and variable speeds of action within a sequence.
\begin{equation}\label{eq:temporal_augmentation}
 \{s'_1,s'_2,...,s'_N\} = G_{temp}(\{s_1,s_2,..,s_M\}),
\end{equation}

\boldparagraph{Translation augmentation.}
We employ translation augmentation to deal with noise coming from inaccuracies in 3D pose estimation.
\begin{equation}\label{eq:trans_augmentation}
  S'_k = G_{trans}(S_k) = S_k + \mathcal{N}(\sigma),
\end{equation}
where $\mathcal{N}(\sigma)$ produces a uniform distribution noise with a standard deviation of $\sigma$.

\boldparagraph{Flipping.}
As our body is mirror-symmetric, we propose a flipping strategy. The flipping function $G_{flip}(\cdot)$ flips left body joints to right in the spatial coordinates and vice versa.

\boldparagraph{Angle augmentation.}
To perform data augmentation, on joint angles we compute
\begin{equation}\label{eq:angle}
  \Theta'_k = G_{angle}(\Theta_k) = \Theta_k +\mathcal{MN}(C),
\end{equation}
$\mathcal{MN}(C)$ denotes a multivariate normal distribution with covariance matrix $C$ which contains high correlations along the diagonal, as explained above.

\boldparagraph{VPoser Augmentation.}
VPoser~\cite{Pavlakos2019SMPL-X} presents a method to learn an embedding space of plausible human poses. We leverage this latent space to further generate matching pairs of skeletal sequences by data augmentation. To this end, we map our pose with VPoser into the latent space and sample nearby location in the latent space. The augmented latent space is then decoded back to the human pose. 
\begin{equation}\label{eq:vposer}
  \Theta'_k = G_{vposer}(\Theta_k)  = V_{dec}(V_{enc}(\Theta_k) +\mathcal{MN}(C)),
\end{equation}
Here, we use the same distribution, $\mathcal{MN}$, for angle augmentation. $V_{enc}(\cdot)$ and $V_{dec}(\cdot)$ correspond to the encoder and  decoder of VPoser, respectively.

\subsection{Implementation Details}\label{sec:implementation_details}
To be robust to  different skeleton sizes, we scale the bone length between the chest joint and pelvis joint to the unit length and resize all the other limb lengths accordingly. We set the chest as the origin of our coordinate system to normalize for translation. We align the bone between the chest and pelvis to the z-axis and the bone between the chest and right shoulder to the y-axis, to account for variations in rotation. We perform a similar normalization for hand skeletons.

We rely on the TCC~\cite{dwibedi2019temporal} code to reproduce their results for the experiments on the H2O dataset and pose-based alignment, following the same hyperparameters, described in~\cite{dwibedi2019temporal}. We provide further details for the parameters of our framework in the supplemental material.

%% file: 04_evaluation.tex
\section{Evaluation}\label{sec:evaluation}

In this section, we first describe the datasets and the corresponding evaluation protocols. We then provide a detailed analysis of our approach, CASA, and compare our approach against the state-of-the-art methods.

\subsection{Datasets}
We verify our model on Penn Action~\cite{Zhang_2013_Penn}, IKEA ASM~\cite{Ben-shabat_2020_Ikea}, and H2O~\cite{Kwon_2021_H2O} datasets. Penn Action is a sports activity dataset. Following previous work~\cite{dwibedi2019temporal,haresh2021learning}, we use the subset of $13$ activities for evaluation. We precisely follow earlier work~\cite{dwibedi2019temporal,haresh2021learning} for training and test splits. IKEA ASM~\cite{Ben-shabat_2020_Ikea} dataset consists of $371$ videos that demonstrate the assembly of four different furniture types. Similarly with LAV~\cite{haresh2021learning}, we conduct our experiments using {\it Kallax\_Drawer\_Shelf} assembly videos (61 for training and 29 for validation). H2O~\cite{Kwon_2021_H2O} is a recent egocentric action recognition and hand-object interaction dataset that provides ground-truth 3D poses for left \& right hands and 6D object poses, along with interaction labels. On this dataset, we select video sequences from the activity, \emph{pouring milk}, which contains monotonic sub-actions. Among 10 subjects performing the action, we select 7 for the training set (27 videos) and 3 for the validation set (11 videos). The sequences have up to 865 frames, and we annotate 10 different phases based on the original action labels, which are only used for the evaluation purposes. %
We will make these new labels for sequence alignment publicly available.
While we use the full-body pose as our input modality in Penn Action and IKEA ASM datasets, we use hand pose as input for the H2O dataset. Particularly for the H2O dataset, our method demonstrates an application of skeletal alignment for hands from an egocentric view, which is highly relevant for augmented reality scenarios. Since the Penn Action dataset does not provide 3D human poses, we estimate the 3D joints of the body using a state-of-the-art body pose estimator~\cite{joo2020eft,rong2021frankmocap}.

\subsection{Evaluation Metrics}
Following literature~\cite{dwibedi2019temporal,haresh2021learning}, we use three different metrics for our evaluation. We first train our network on the training set without using any labels and then evaluate the performance of our approach using the trained embeddings.

\myparagraph{Phase Classification Accuracy} is the per-frame classification accuracy for fine-grained action recognition. To evaluate this metric, we train an SVM classifier on a limited subset of the training data to predict phase labels. 

\myparagraph{Phase Progression} measures how well the {\it progress} of a process or action is captured by the embeddings. We follow previous work~\cite{dwibedi2019temporal} to use a linear regressor on the embeddings to predict the phase progression values. It is computed as the average $R-$squared measure, given by
\begin{align}
    R^{2} = 1 - \frac{\sum_{i=1}^{n}(y_{i}-\hat{y_{i}})^{2}}{\sum_{i=1}^{n}(y_{i}-\bar{y_{i}})^{2}} \;,  \label{eq:r_square}   
\end{align}
where $y_{i}$ is the ground truth phase progress value, $\bar{y}$ is the mean of all $y_{i}$ and $\hat{y_{i}}$ is the prediction made by the linear regression model. The maximum value of this measure is 1.

\myparagraph{Kendall's Tau~\cite{dwibedi2019temporal}} is a statistical measure that can determine how well-aligned two sequences are in time. It is in the range of $[-1,1]$ where
a value of $1$ implies that the videos are perfectly aligned, while a value of $-1$ implies that the videos are aligned in reverse order.  Since this metric assumes a strictly monotonic order of the actions, it is evaluated only on the Penn Action dataset.

\subsection{Comparison to the State-of-the-Art}

\begin{table}
\begin{center}
\begin{adjustbox}{width=1.0\columnwidth,center}
\begin{tabular}{|c|c|c|c|c|c|c|}
\hline
\multirow{2}{*}{Dataset} & \multirow{2}{*}{Method} & \multirow{2}{*}{Pose} & ImageNet & \multicolumn{3}{c|}{\% of Labels $\rightarrow$}\\
\cline{5-7}
& & & pre-trained& 0.1 &0.5&1.0\\ 
 \hline
\multirow{7}{*}{Penn Action~\cite{Zhang_2013_Penn}} 
& SaL~\cite{misra2016shuffle} & $\cdot$ & \checkmark& 74.87 &78.26&79.96	\\
& TCN~\cite{sermanet2018time} & $\cdot$ & \checkmark& 81.99 &83.67&84.04	\\
& TCC~\cite{dwibedi2019temporal} & $\cdot$ & \checkmark & 79.72 & 81.11 & 81.35 \\
& LAV~\cite{haresh2021learning} & $\cdot$ &  \checkmark& 83.56 & 83.95 & 84.25 \\
& TCC~\cite{dwibedi2019temporal} & \checkmark & $\cdot$ & 79.53 & 83.75 & 84.51 \\
& LAV~\cite{haresh2021learning} & \checkmark & $\cdot$ & 79.83 & 80.20 & 80.20 \\
& CASA (ours) & \checkmark  & $\cdot$& \textbf{88.55} & \textbf{91.87} & \textbf{92.20}\\\hline
\multirow{5}{*}{IKEA ASM~\cite{Ben-shabat_2020_Ikea}}
& TCC~\cite{dwibedi2019temporal} & $\cdot$ &  \checkmark& 27.74&  25.70&  26.80  \\
&LAV~\cite{haresh2021learning} & $\cdot$  &  \checkmark& \textbf{29.78} & 29.85 & 30.43\\
& TCC~\cite{dwibedi2019temporal} & \checkmark & $\cdot$ & 11.95 & 13.53 & 18.60 \\
& LAV~\cite{haresh2021learning} & \checkmark & $\cdot$ & 14.52 & 16.31 & 18.63 \\
& CASA (ours) & \checkmark  & $\cdot$ & 21.32 & \textbf{31.52} & \textbf{31.06} \\\hline 
\multirow{5}{*}{H2O~\cite{Kwon_2021_H2O}}
& TCC~\cite{dwibedi2019temporal} & $\cdot$ &  \checkmark & 43.30 & 52.48 & 52.78 \\
& LAV~\cite{haresh2021learning} & $\cdot$ &  \checkmark& 23.48  & 36.41  & 36.38\\
& TCC~\cite{dwibedi2019temporal} & \checkmark & $\cdot$ & 30.40 & 40.20 & 42.70 \\
& LAV~\cite{haresh2021learning} & \checkmark & $\cdot$ & 37.05 & 39.50 & 40.45 \\
& CASA (ours) & \checkmark & $\cdot$ & \textbf{43.50} & \textbf{62.51} & \textbf{68.78} \\
\hline
\end{tabular}
\end{adjustbox}
\end{center}
\vspace{-4mm}
\caption{\small {\bf Phase classification results.} We compare our phase classification accuracy to those of both RGB and pose based methods on three different datasets. Our method produces the state-of-the-art results in most cases.}
\vspace{-4mm}
\label{table:phase_classification}
\end{table}

We compare our self-supervised skeletal sequence learning approach against several different approaches~\cite{dwibedi2019temporal,haresh2021learning,misra2016shuffle,sermanet2018time,sun2020view}, including the recent self-supervised video representation learning techniques, TCC~\cite{dwibedi2019temporal}, and LAV~\cite{haresh2021learning}, that use alignment as a pretext task. Previous approaches do not report results using pose data as input. Therefore we reproduce the results of these baselines to be able to benchmark our results against them by following the implementation details of~\cite{dwibedi2019temporal,haresh2021learning}.
Using precisely the same feature extractor for processing poses, we compare our approach against them. For feature extraction, we use two non-linear fully connected layers which have the same dimension with our input to keep the same amount of information. We did our best to make fair comparison by following the same hyperparameters from LAV~\cite{haresh2021learning} and TCC~\cite{dwibedi2019temporal} 
except for the learning rate, which we set as 0.00005 for the image, and 0.0005 for pose, as we observed better convergence with these learning rates for different input modalities. %

We compare our phase classification accuracy to the state-of-the-art~\cite{dwibedi2019temporal,haresh2021learning,misra2016shuffle,sermanet2018time} in Table~\ref{table:phase_classification}. We significantly outperform the existing approaches for all datasets and for all the fractions of labels that are used to train the classifier, except for the case of training with 10\% of the labels in the IKEA ASM dataset. Limited performance for 10\% of the labels in the IKEA ASM dataset is due to the noisy pose estimates on this dataset resulting from object occlusions, differences in viewpoints (\eg sitting vs standing during furniture assembly), and missing hand poses that provide informative cues for the assembly task. For TCC~\cite{dwibedi2019temporal} and LAV~\cite{haresh2021learning}, the pose input results in lower accuracy than the image input on the IKEA ASM dataset due to missing contextual information related to object interactions. Yet, our approach achieves better overall accuracy than the existing approaches that either use image or pose as input on this dataset. Our method reasons about fine-grained actions, both, by accounting for contextual information through our transformer-based self- and cross-attention mechanism, and, by exploiting 3D poses, which provide a detailed understanding of subtle human motions.

\begin{table}
\begin{center}
\begin{adjustbox}{width=0.9\columnwidth,center}
\begin{tabular}{|c|c|c|c|c|}
\hline
Method& Pose  & \shortstack{ImageNet\\  pre-trained}& Progress  & $\tau$ \\
\hline
TCN~\cite{sermanet2018time} & $\cdot$ & \checkmark & 0.6762 &0.7328	\\
SaL~\cite{misra2016shuffle} & $\cdot$ & \checkmark & 0.5943 &0.6336	\\
Pr-VIPE~\cite{sun2020view} & \checkmark* & $\cdot$ & $\cdot$ &	0.7476\\
TCC~\cite{dwibedi2019temporal} & $\cdot$ & $\cdot$  & 0.4304 & 0.4529 \\
LAV~\cite{haresh2021learning} & $\cdot$  & $\cdot$ & 0.3853 & 0.4929 \\
TCC~\cite{dwibedi2019temporal} & $\cdot$ & \checkmark & 0.6638 & 0.7012 \\
LAV~\cite{haresh2021learning} & $\cdot$ & \checkmark  & 0.6613 & 0.8047 \\
Hadji~\cite{hadji2021representation} & $\cdot$ & \checkmark & $\cdot$ &	0.7829\\
TCC~\cite{dwibedi2019temporal} & \checkmark& $\cdot$  & 0.6268 & 0.6267 \\
LAV~\cite{haresh2021learning} & \checkmark & $\cdot$  & 0.6404 & 0.6983 \\
CASA (ours) & \checkmark & $\cdot$  & \textbf{0.9449} & \textbf{0.9728}\\\hline
\end{tabular}
\end{adjustbox}
\end{center}
\vspace{-4mm}
\caption{\small{\bf Video progress and Kendall's tau results.} We compare our method to other RGB and pose based methods. Note that * uses 2D poses. Our method achieves the best results on the Penn Action dataset.}
\vspace{-4mm}
\label{table:progress}
\end{table}

In Table~\ref{table:progress}, we further report our phase progression and Kendall's tau results as compared to the state-of-the-art. Remember that these metrics respectively measure how well the progress of an action is and how well aligned two sequences are in time. Our approach outperforms earlier approaches on these metrics by a large margin ($0.27$ improvement on phase progression and $0.17$ improvement on Kendall's tau). We attribute this to the fact our method exploits positional encodings to encode temporal frame location which is a valuable cue for understanding the progress and alignment of actions. Using an attention-based architecture, our method gathers contextual information from the whole sequence during alignment which results in superior accuracy than previous approaches that rely on \emph{only} local context. We provide further quantitative results of our approach for online sequence alignment in our supplemental material.

\subsection{Ablation Studies}  \label{sec:ablation}

In Table~\ref{table:ablation_network}, we provide an ablation study on the Penn Action dataset to analyze the influence of different network components. All our design choices consistently improve our overall accuracy. The improvement is particularly pronounced for positional encodings and attention layers. While positional encodings provide local information about frame location, attention layers help gather contextual information within the same sequence and across two sequences. The projection head also results in a considerable improvement in accuracy for all the metrics, showing the importance of the nonlinear mapping before applying a self-supervised loss, in line with the recent literature on self-supervised learning~\cite{chen2020simple}.

We present further ablation studies on the influence of different types of data augmentation strategies in Table~\ref{table:ablation_4d}. All the augmentation strategies, combined together, results in consistently high accuracies for all the metrics. While the temporal augmentation results in about $2\%$ increase in the phase classification accuracy, 3D spatial augmentation brings in another $1\%$ improvement in phase classification and Kendall's Tau, which demonstrates the individual contributions and complementary nature of different strategies.

\begin{table}
\vspace{-4mm}
\begin{center}
\begin{adjustbox}{width=1.0\columnwidth,center}
\begin{tabular}{|c|c|c|c|}
\hline
Method & Classification(\%)& Progress & $\tau$ \\
 \hline
w/o positional encoding & 69.01 & 0.3361 & 0.3415 \\
w/o projection head & 89.87 & 0.8852 & 0.9713 \\
w/o self attention layers & 91.24 & 0.9193 & 0.9310 \\
w/o cross attention layers  & 92.04 & 0.9316 & 0.9616 \\
All & \textbf{92.20} & \textbf{0.9449} & \textbf{0.9728} \\
\hline
\end{tabular}
\end{adjustbox}
\end{center}
\vspace{-4mm}
\caption{\small {\bf Influence of different components of our model.} We ablate on the Penn Action dataset to analyze our different design choices.}
\label{table:ablation_network}
\end{table}

\begin{table}
\begin{center}
\begin{adjustbox}{width=1.0\columnwidth,center}
\begin{tabular}{|c|c|c|c|}
\hline
Method & Classification(\%)& Progress & $\tau$ \\
 \hline
No Aug. & 89.95 & 0.8729 & 0.9653 \\
Temp. Aug. & 91.78&  0.9446  & 0.9621  \\
w/o Ang. & 92.75  & 0.9397 & 0.9719  \\
w/o Trans.  & 92.64  & 0.9338 & 0.9722 \\
w/o Vposer  & 92.64  & 0.9379  & 0.9710 \\
w/o Flip  & \textbf{92.94} & 0.9414  & 0.9710 \\
All & 92.20 & \textbf{0.9449} & \textbf{0.9728} \\
\hline
\end{tabular}
\end{adjustbox}
\end{center}
\vspace{-4mm}
\caption{\small {\bf Ablation study for 4D augmentation.} The best result is depicted in bold. We ablate on the Penn Action dataset to analyze different data augmentation strategies.}
\vspace{-4mm}
\label{table:ablation_4d}
\end{table}

\begin{figure}[t]
\begin{center}
\captionsetup[subfigure]{labelformat=empty}
\vspace{-5mm}
\hspace{-0.1cm}\subfloat[TCC]{\includegraphics[width=0.45\columnwidth]{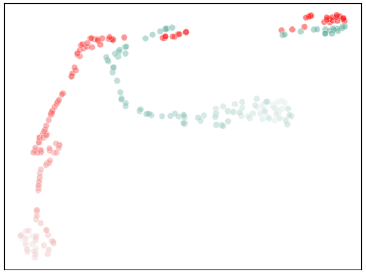}}  
\hspace{-0.0cm}\subfloat[OURS]{\includegraphics[width=0.46\columnwidth]{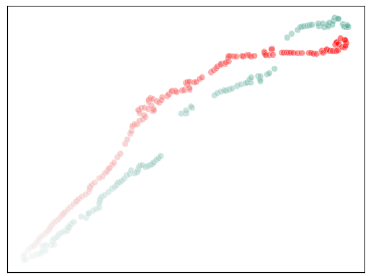}}
\end{center}
\vspace{-0.6cm}
\caption{\small {\bf t-SNE visualization of the embedding space learned by CASA.}
For this visualization, we select two different sequences from {\it baseball\_pitch}. Our method is able to preserve temporal context and align corresponding frames across videos.}
\vspace{-3mm}
\label{fig:tsne}
\end{figure}

\begin{figure}[t]
\begin{center}
\captionsetup[subfigure]{labelformat=empty}
\subfloat[TCC]{\includegraphics[width=0.218\textwidth]{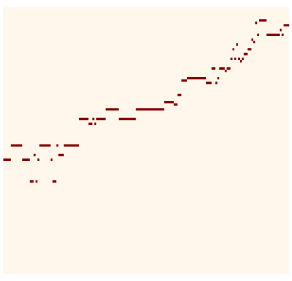}} \hspace{2mm}
\subfloat[CASA]{\includegraphics[width=0.22\textwidth]{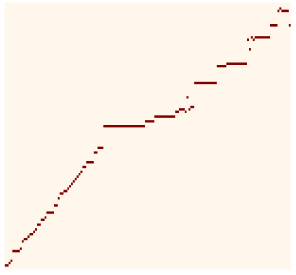}}\\
\vspace{-6mm}
\end{center}
\caption{\small {\bf Alignment between two sequences.} The x-axis is the time frame of the source sequence and the y-axis is the time frame of the target sequence. %
We show the closest matching frames across source and target sequences. For visualization, we select two different sequences from {\it baseball\_pitch}. We observe that CASA preserves the temporal context and results in smoother alignments.}
\vspace{-4mm}
\label{fig:dist_matrix}
\end{figure}

We present the t-SNE embeddings~\cite{van2008visualizing} of the representation learned by CASA on two different sequences in Fig.~\ref{fig:tsne}. Color scale demonstrates the corresponding time frames of a sequence, from start to end. We demonstrate that our approach learns a smooth representation, in which temporally close frames are mapped to nearby positions in the embedding space. Furthermore, the corresponding frames across the two videos are embedded in similar locations. This structure of the embedding space demonstrates the potential and reliability of our method for sequence alignment. We show qualitative examples of alignment between two sequences in Fig.~\ref{fig:teaser}. More qualitative results can be found in our supplemental material. We further show frame-wise matches across two sequences, in comparison to TCC, in Fig.~\ref{fig:dist_matrix}. We observe that CASA preserves the temporal context and results in smoother alignments. %

%% file: 05_conclusion.tex
\section{Conclusion}\label{sec:conclusion}
In this paper, we propose a self-supervised learning framework that uses skeletal sequence alignment as a proxy task. The proposed CASA approach uses the self and cross attention layers in Transformers to transform the local features to be context- and position-dependent, which is crucial for CASA to obtain high-quality sequence alignments. We further propose to augment the skeletal sequences in 3D space and time to generate examples for matching and training a self-supervised loss to minimize alignment score across sequences. Our experiments show that CASA achieves state-of-the-art performances on phase action classification, phase progression, and Kendall's tau scores on multiple datasets. 

Our method, CASA, relies on off-the-shelf pose estimators to compute human pose, which is used as an input to our framework for alignment. Wrong predictions of the off-the-shelf pose estimator will result in inaccuracies in sequence alignment, which is a limitation of our approach. End-to-end learning from RGB images for skeletal alignment using a pretrained pose estimator would be an interesting future direction to overcome this limitation.

\boldparagraph{Acknowledgements.} Taein Kwon was supported by the Microsoft MR \& AI Z\"urich Lab PhD scholarship. 
The authors thank Jonas Hein, Mihai Dusmanu, Paul-Edouard Sarlin, Luca Cavalli, Yao Feng, and Weizhe Liu for helpful discussions.